\documentclass[conference]{IEEEtran}
\IEEEoverridecommandlockouts
\usepackage{cite}
\usepackage{amsmath,amssymb,amsfonts}
\usepackage{algorithmic}
\usepackage{graphicx}
\usepackage{textcomp}
\usepackage{xcolor}
\usepackage{threeparttable}
\usepackage{multirow}
\def\BibTeX{{\rm B\kern-.05em{\sc i\kern-.025em b}\kern-.08em
    T\kern-.1667em\lower.7ex\hbox{E}\kern-.125emX}}
\begin{document}

\title{SAMV-DUSt3R: Instance-Centric 3D Scene Decoupling from Sparse Multi-Views\\
\thanks{\textsuperscript{*}Equal Contribution, \textsuperscript{†}Corresponding author.}
}

\author{\IEEEauthorblockN{Langxu Zhao, Zuan Gu\textsuperscript{*}, Yingdan Zhang\textsuperscript{*},Pengfei Zhao\textsuperscript{*}, Tianhan Gao\textsuperscript{†}}
\IEEEauthorblockA{\textit{Northeastern University} Shenyang, China}
}
\maketitle

\begin{figure*}[t]
\centering
\includegraphics[width=0.95\textwidth]{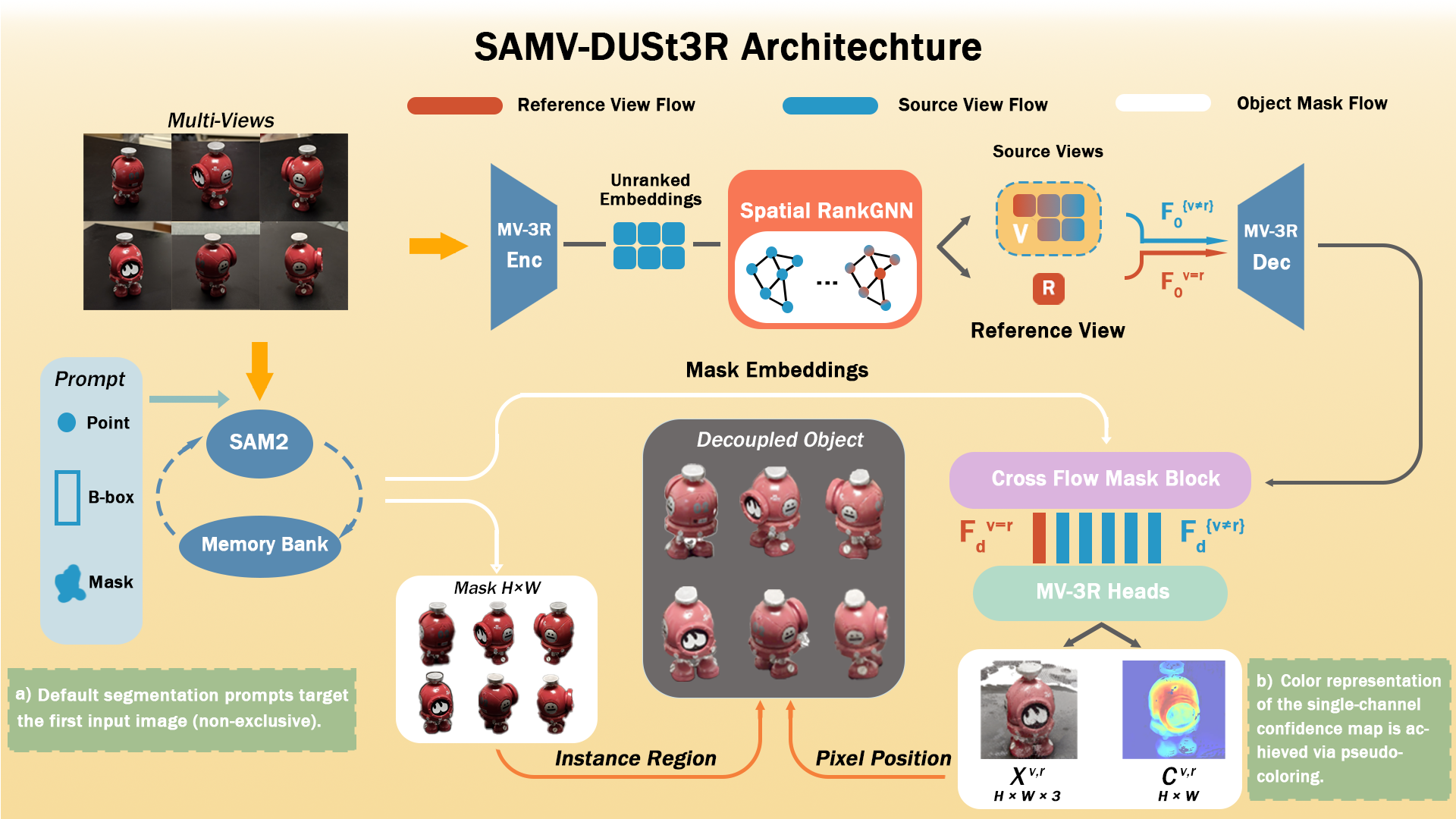} 
\caption{We present \textbf{SAMV-DUSt3R}, a 3D object decoupling model based on MV-DUSt3R and SAM2. Given a set of unordered and uncalibrated RGB views along with user-specified prompts indicating the target object for decoupling, the model outputs the decoupled 3D object model. The figure shows that \textbf{Spatial RankGNN} ranks the spatial information provided by each view to select the reference view R and the features interact with multi-view object masks in \textbf{Cross Flow Mask Block}, thereby enhancing the confidence level within the target region. This process ultimately yields the decoupled 3D object model extracted from the input views.}
\vspace{-15pt}
\label{fig:overview}
\end{figure*}

\begin{abstract}
With the rising demand to decouple objects from 3D scenes, we propose SAMV-DUSt3R, an end-to-end model that injects SAM2 2D masks into MV-DUSt3R reconstruction. A Cross Flow Mask Block uses these masks to steer the network toward the target instance, jointly improving shape accuracy and achieving object-level disentanglement without multi-stage pipelines. To ensure reconstruction stability, a lightweight Spatial RankGNN selects the optimal reference view with a selection accuracy of 73.5\%. Extensive experiments demonstrate that our method boosts average reconstruction precision by 11\% across various metrics compared to state-of-the-art baselines. These results reveal a strong instance-disentanglement capability and clear benefits for driving, robotics, AR/VR, and heritage digitisation.
\end{abstract}


\section{Introduction}
Achieving interactive and controllable 3D decoupling \cite{DBLP:conf/eccv/PengWYLC22,10219827,DBLP:conf/aaai/Lin0HTLLWSY25} is made possible via reconstruction . Models based on DUSt3R  \cite{wang2024dust3rgeometric3dvision,tang2024mvdust3rsinglestagescenereconstruction,DBLP:conf/eccv/LeroyCR24,DBLP:journals/corr/abs-2410-03825,DBLP:journals/corr/abs-2411-16877}have garnered significant attention, which products pointmaps as 3D representation matching the dimensions of the source views and eliminates the dependency on camera parameters inherent in traditional Multi-View Stereo \cite{10.1007/978-3-319-46487-9_31,wang2020patchmatchnetlearnedmultiviewpatchmatch,zhang2020visibility} methods. Researchers proposed MV-DUSt3R \cite{tang2024mvdust3rsinglestagescenereconstruction}, which employs multi-view decoder blocks to simultaneously process an arbitrary number of views in one forward pass, avoiding cascaded optimization and significantly improving both reconstruction quality and speed. Nevertheless, MV-DUSt3R relies on a single reference view, potentially degrading reconstruction quality in regions with large viewing angles. Furthermore, current DUSt3R-based methods focus on holistic scene reconstruction and cannot directly obtain, or leverage input prompts to obtain, the 3D representation of user-specified targets within the scene. To achieve such capability, we propose the SAMV-DUSt3R model. It combines the SAM2 \cite{ravi2024sam2segmentimages} and MV-DUSt3R, leveraging SAM2's promptability, generalization capability, and powerful segmentation performance to provide MV-DUSt3R with instance-level reconstruction potential. This enables 3D object disentanglement from sparse views. Our method designs a Cross Flow Mask Block to fuse the segmentation masks obtained from SAM2 with the decoded view information derived by MV-DUSt3R. This allows the segmentation masks of target regions to directly participate in the reconstruction process. Concurrently, addressing the reconstruction instability induced by reference view choices, the covisibility graphs from SfM \cite{7780814,Hartley_Zisserman_2004} outputs motivate our design of Spatial RankGNN to deliver near-instantaneous view selection at negligible computational cost. Rigorous experimentation demonstrates that SAMV-DUSt3R can achieve 3D scene reconstruction from sparse views centered around specific targets, along with user-specified object disentanglement. In summary, our contributions include: \textbf{(1) SAMV-DUSt3R}: A 3D decoupling model generates the 3D reconstruction result of one target from sparse RGB views by prompts, without requiring camera parameters or image calibration. \textbf{(2) Cross Flow Mask Block}: Fuses the segmentation masks of the target to be disentangled with the decoded view information from MV-DUSt3R, enabling the model to incorporate object-level information. \textbf{(3) Spatial RankGNN}: A lightweight GNN architecture trained on geometric patterns within SfM results for intelligent reference view selection to achieve maximal reconstruction accuracy.

\section{Related Works}
\textbf{Feed-forward Reconstruction.} DUSt3R bypasses traditional pipelines by regressing 3D coordinates directly via a Transformer encoder-decoder without camera parameters. MASt3R \cite{DBLP:conf/eccv/LeroyCR24} enhances this with dense feature matching for improved keypoint correspondence. To eliminate global optimization, Spann3R \cite{wang20243d} adopts an incremental approach using a Spatial Memory structure. MV-DUSt3R \cite{tang2024mvdust3rsinglestagescenereconstruction} addresses depth inaccuracies and optimization costs in sparse views through a Multi-View Decoder and Cross-Reference-View Attention, supporting flexible view counts and ordering. Additionally, Fast3R \cite{Yang_2025_Fast3R} enables parallel processing for up to 1500 images in a single forward pass. Beyond point-based methods, InstantSplat \cite{fan2024instantsplat} leverages DUSt3R’s geometric priors to initialize and jointly optimize 3D Gaussian Splatting \cite{kerbl3Dgaussians} and camera parameters.

\textbf{3D Decoupling.} 3D decoupling disentangles attributes like shape, material, and texture for independent manipulation. DecoupledGaussian \cite{decoupledGaussian} integrates 3D Gaussian Splatting with planar constraints and a Joint Poisson Field to reconstruct precise, complete 3D models. Whole-component segmentation also facilitates decoupling. SAMPart3D \cite{yang2024sampart3d} projects DINOv2-extracted 2D features into 3D space and utilizes a Scale-Conditioned MLP to distill multi-granularity SAM masks for object-part disentanglement. For mesh-based decoupling, SAMesh \cite{tang2025segmentmesh} processes multi-view renders and geometric modalities (e.g., surface normals) via SAM, then employs a match graph and community detection to achieve 3D mesh segmentation.

\section{Method}
We develop SAMV-DUSt3R to decouple prompt-specified 3D objects within scenes from sparse RGB views. Consistent with MV-DUSt3R's outputs of $Head^{ref}$ and $Head^{src}$, SAMV-DUSt3R leverages predicted 3D Pointmaps $\textbf{X}^{v,r}\in\mathbb{R}^{H \times W \times 3}$ and Confidencemaps $\textbf{C}^{v,r}\in\mathbb{R}^{H \times W}$ to reconstruct 3D targets. Our contributions include training Spatial RankGNN using covisibility graphs from SfM pipelines to perform reference-view ranking. Moreover, we design Cross Flow Mask Block to integrate segmentation mask with decoder features from the reconstruction backbone network and introduce Masked Confidence Loss that lifts predicted confidence for target objects by characterizing the confidence distribution within segmentation regions on each \textbf{C} map. 

\subsection{Architecture Design}
We eliminate manual selection of reference views from sparse input RGB images. All images are processed through MV-DUSt3R’s Enc to obtain image embeddings, which are then fed into Spatial RankGNN, which automatically selects the optimal reference view based on its ranking output. The embeddings of the reference view and other views are passed to the Dec, where the decoding logic will be maintained. Concurrently, the multi-view images are processed in parallel by SAM2 to generate corresponding segmentation masks and mask embeddings. The Cross Flow Mask Block get the sequence representations and mask embeddings and output the multi-view $Flow_{d}$ into last two Heads. This pipeline yields \textbf{Predicted 3D point map}, \textbf{Confidence map} and \textbf{Segmentation mask} as a triplet \textbf{(X, C, M)} per input view.

\textbf{Spatial RankGNN.} Graph Neural Networks have unique application areas in 3D vision. However, in recent 3D reconstruction research, GNNs are unsuitable for point/pixel-level tasks due to their requirement for iterative message passing between nodes, which results in high computational complexity in large-scale point cloud or mesh scenes. For MV-DUSt3R’s reference view selection problem, we must consider both how to balance stereo information transfer across multiple images and ensure that the introduced method does not excessively impact model inference efficiency. Inspired by researchs on GNNs for multi-view camera relocalization \cite{Xue2020LearningMC}, we approach reference view selection from graph structure: The stereo information exchange process between images can be modeled as a graph structure, where each view serves as a node, the transmission process forms edges, and the quantity of provided information acts as edge weights. Nodes that provide more information are often more suitable as reference views. We design the \textbf{RankMessagePassing} and the \textbf{Spatial RankGNN} module which stacks multiple layers of this unit followed by a global readout. The RankMessagePassing module implements a graph node message-passing mechanism based on multi-head self-attention. It computes scaled attention scores between all node pairs, normalizes them via softmax, and aggregates weighted value vectors to form messages. These multi-head outputs are unified through another 1×1 convolution. A residual update is then performed by concatenating the original features with the messages, followed by feature fusion through a 1×1 convolution with ReLU activation. Layer normalization is finally applied to stabilize feature representations. The Spatial RankGNN architecture models image spatial positions as graph nodes. Input features are initially projected to a hidden dimension. Batch and spatial dimensions are flattened into a unified node sequence. Multiple stacked RankMessagePassing layers enable global node interactions, permitting cross-image attention where each spatial pixel participates as a node in multi-head attention computations. Spatial mean pooling derives image-level embeddings, which are transformed into normalized scores through a two-layer MLP with ReLU and Sigmoid activations, facilitating end-to-end image ranking prediction.
In contrast to conventional GNNs that iteratively perform multiple rounds of message passing between adjacent nodes, RankMessagePassing first employs convolutions to generate Q, K, V matrices in parallel. Subsequently, a single batched matrix multiplication followed by a softmax operation facilitates global information exchange among all node pairs. This entire process constitutes a global attention mechanism with \textbf{$\mathcal{O}(N^2)$}, where N denotes the number of input views. By stacking merely L layers, long-range information propagation is achieved, resulting in \textbf{$\mathcal{O}(LN^2)$}.
\begin{figure}[t]
\centering
\includegraphics[width=0.45\textwidth]{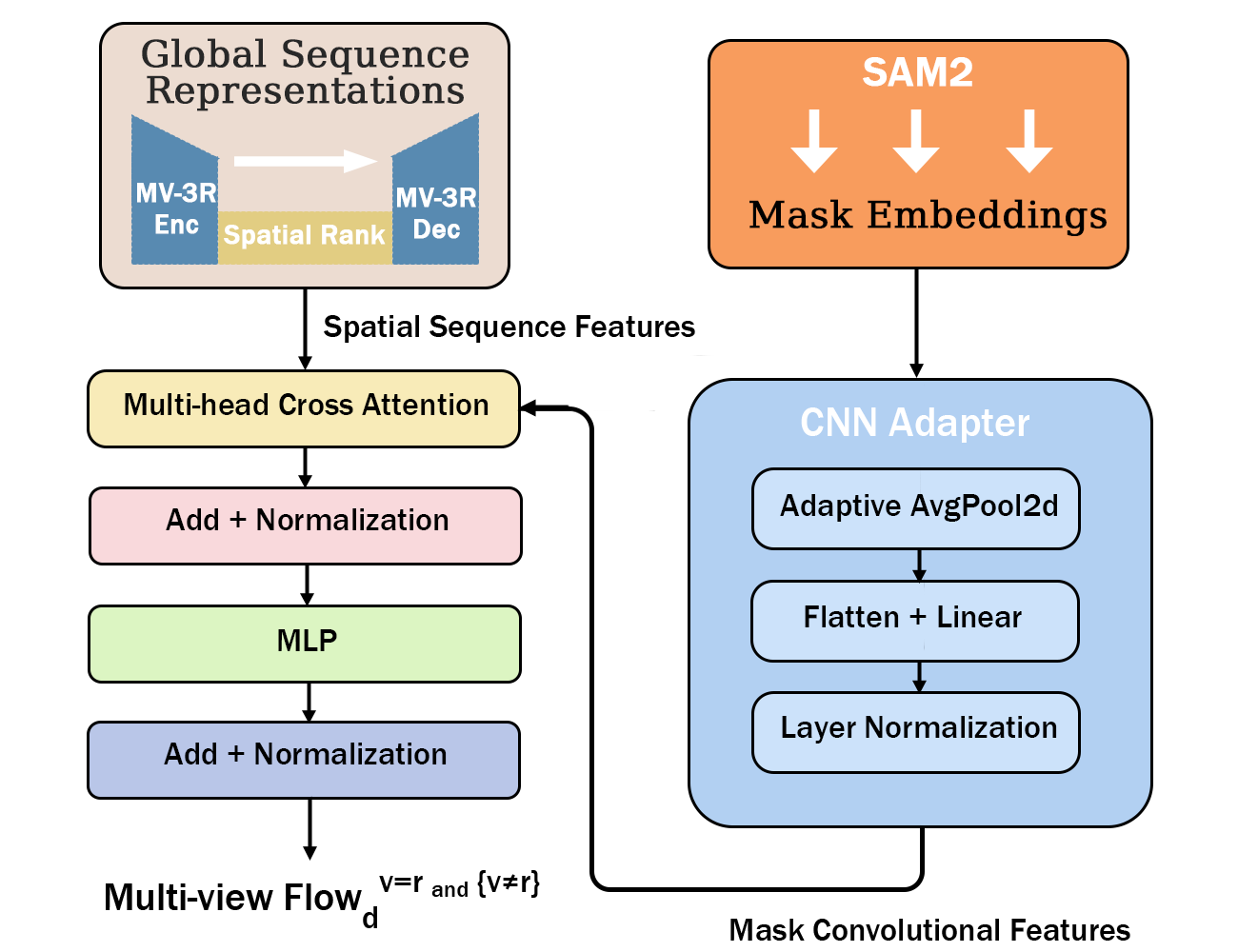} 
\caption{The \textbf{Cross Flow Mask Block} is designed to infuse mask-derived convolutional features into spatial sequence representations from \textbf{Enc-Dec}. }
\vspace{-15pt}
\label{fig:cfmb}
\end{figure}

\begin{figure*}[t]
\centering
\includegraphics[width=0.95\textwidth]{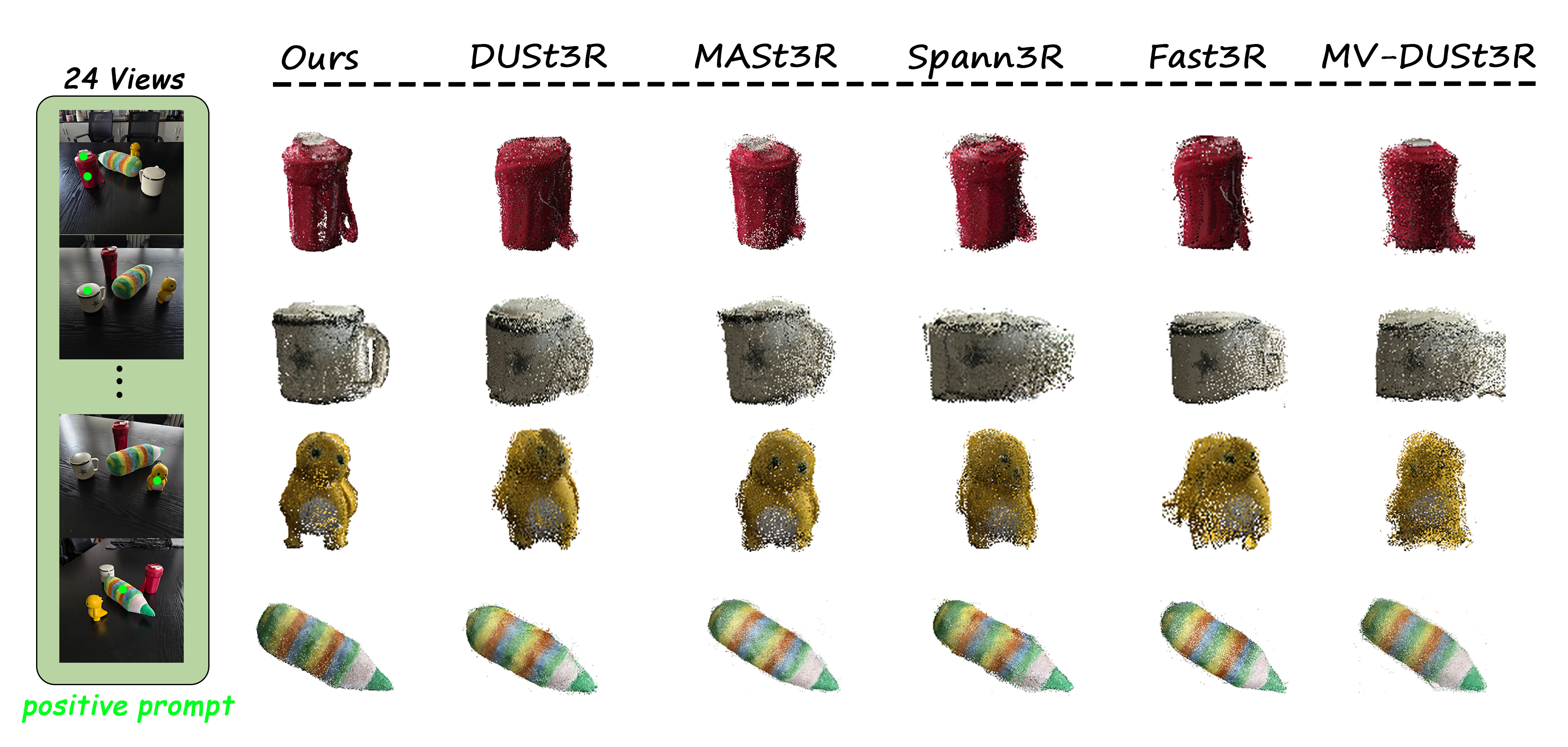} 
\caption{To demonstrate SAMV-DUSt3R's performance in real-world, we captured 24 uncalibrated and unordered views surrounding four objects. Using point prompts, we generated segmentation masks which were subsequently applied to the outputs of other 3R methods on NVIDIA A800 GPU. The figure presents the decoupled 3D object pointclouds reconstructed by each method. The results are visualized from a consistent viewpoint.}
\label{fig:comparison}
\end{figure*}

\textbf{Cross Flow Mask Block.} To enable enhanced spatial guidance for sequence-level features using 2D object masks, we propose the \textbf{Cross Flow Mask Block}, a cross attention module designed to explicitly infuse mask-derived convolutional cues into the sequence representations learned from MV-DUSt3R with Spatial RankGNN. This module integrates two heterogeneous feature sources: a sequence embedding produced by the encoder-decoder pipeline and convolutional features extracted from 2D object masks through a lightweight CNN-based adapter. The CNN adapter transforms the mask features into a token-aligned representation via spatial pooling, preparing them to serve as keys and values in a multi-head cross-attention mechanism. Within this block, the MV-DUSt3R sequence tokens are used as queries to attend to the spatial mask features, thereby enabling fine-grained spatial alignment between 3D scene understanding and explicitly localized 2D mask priors. The attention-enhanced sequence is further refined via residual connections, layer normalization, and a position-wise feedforward network to preserve representational fidelity and facilitate gradient flow. Overall, the Cross Flow Mask Block acts as a mask-aware attention filter, guiding global sequence representations to focus on object regions in a differentiable and data-driven manner. This design effectively bridges global context and localized spatial priors.
\begin{table*}[t]
\caption{Metrics in MVS Reconstruction and NVS}
\centering
\begin{threeparttable}
\begin{tabular}{ c | c | c c c | c c | c c c}
\hline
\multirow{1}{*}{ }
& \multirow{1}{*}{Method}
& \multirow{1}{*}{ND$\downarrow$}
& \multirow{1}{*}{DAc$\uparrow$}
& \multirow{1}{*}{CD$\downarrow$}
& \multirow{1}{*}{Acc$\downarrow$}
& \multirow{1}{*}{Comp$\downarrow$}
& \multirow{1}{*}{PSNR/P$_{\text{mask}}$$\uparrow$}
& \multirow{1}{*}{SSIM/S$_{\text{mask}}$$\uparrow$}
& \multirow{1}{*}{LPIPS/L$_{\text{mask}}$$\downarrow$} \\

\hline
\multirow{8}{*}{\rotatebox{90}{4 views}}
& DUSt3R$_{\text{512}}$        & 8.9  & 5.0 & 3.1 & 8.6 & 8.6 & \textbf{11.2} / 20.9 &3.4 / \textbf{9.1} & \textbf{5.9} / 6.2   \\
& MASt3R     & 8.5  & 5.2 & 3.1 & 10.0 & 9.8 & 9.9 / 19.9 &2.8 / 9.0 &6.1 / 8.6   \\
& Spann3R$_{\text{1.01}}$        & 6.9  & 6.3 & 2.9 & 8.8 & 8.3 & \textbf{11.2} / \textbf{21.2} & \textbf{3.6} / \textbf{9.1} & \textbf{5.9} / 5.9  \\
& Fast3R    & 7.2  & 6.1 & 2.7 & 8.7 & 8.3 & 11.1 / 21.1 &3.5 / \textbf{9.1} & \textbf{5.9} / \textbf{5.3}   \\
& MV-DUSt3R   & 8.3  & 5.4 & 4.3 & 9.9 & 9.4 & 11.0 / 20.9 &3.3 / \textbf{9.1} &6.0 / 5.8   \\
& MV-DUSt3R$_{\text{object}}$   & 8.0  & 5.0 & 4.3 & 9.8 & 9.0 & \textbf{11.2} / 20.9 &3.4 / \textbf{9.1} &6.0 / 5.6   \\
& SAMV-DUSt3R$_{\text{unrank}}$   & 6.9  & 6.1 & 2.6 & 7.6 & 7.5 & \textbf{11.2} / 21.1 &3.5 / \textbf{9.1} &6.1 / 5.5   \\
& Ours           & \textbf{6.6}  & \textbf{6.6} & \textbf{2.4} & \textbf{6.2} & \textbf{6.3} & \textbf{11.2} / \textbf{21.2} &3.5 / \textbf{9.1} & \textbf{5.9} / \textbf{5.3}   \\
\hline
\hline
\multirow{8}{*}{\rotatebox{90}{12 views}}
& DUSt3R$_{\text{512}}$        & 8.7  & 5.2 & 2.7 & 8.6 & 8.5 & 13.3 / 26.3 & \textbf{4.9} / \textbf{9.5} & \textbf{4.8} / \textbf{2.5}   \\
& MASt3R     & 8.7  & 5.3 & 2.6 & 10.0 & 9.8 & \textbf{13.4} / 23.8 &4.7 / 9.3 &5.1 / 3.8   \\
& Spann3R$_{\text{1.01}}$        & 6.9  & 6.3 & 2.4 & 8.9 & 8.4 & 13.2 / 25.4 &4.7 / 9.4 &5.0 / 3.2   \\
& Fast3R    & 7.2  & 6.1 & 2.3 & 8.4 & 8.1 & 12.9 / 24.5 &4.6 / 9.4 &5.0 / 3.1   \\
& MV-DUSt3R   & 8.0  & 5.7 & 3.8 & 9.8 & 9.3 & 11.9 / 22.1 &4.1 / 9.2 &5.5 / 5.1   \\
& MV-DUSt3R$_{\text{object}}$   & 7.9  & 5.1 & 3.7 & 9.7 & 8.8 & 12.0 / 23.7 &4.1 / 9.2 &5.5 / 4.5   \\
& SAMV-DUSt3R$_{\text{unrank}}$   & 6.9  & 6.2 & 2.4 & 7.0 & 7.1 & 12.5 / 24.0 &4.7 / 9.2 &5.5 / 4.3   \\
& Ours           & \textbf{6.2}  & \textbf{6.8} & \textbf{1.9} & \textbf{5.5} & \textbf{5.8} & 13.2 / \textbf{26.6} & \textbf{4.9} / 9.3 &5.0 / 3.0   \\
\hline
\hline
\multirow{8}{*}{\rotatebox{90}{24 views}}
& DUSt3R$_{\text{512}}$        & 8.5  & 5.3 & 2.6 & 8.6 & 8.5 & 14.4 / 28.1 &5.7 / \textbf{9.6} & \textbf{4.1} / \textbf{1.6}   \\
& MASt3R     & 8.2  & 5.6 & 2.6 & 10.0 & 9.9 & 14.7 / 28.0 &\textbf{5.8} / \textbf{9.6} & \textbf{4.1} / 2.2   \\
& Spann3R$_{\text{1.01}}$        & 6.9  & 6.3 & 2.3 & 8.9 & 8.4 & 14.8 / \textbf{28.5} &5.7 / \textbf{9.6} &4.2 / 1.9  \\
& Fast3R    & 7.2  & 6.1 & 2.2 & 8.3 & 8.1 & 14.4 / 28.3 &5.7 / \textbf{9.6} &\textbf{4.1} / 1.9   \\
& MV-DUSt3R   & 8.0  & 5.6 & 3.7 & 9.8 & 9.3 & 12.9 / 25.4 &5.1 / 9.4 &4.6 / 3.5   \\
& MV-DUSt3R$_{\text{object}}$   & 7.7  & 5.0 & 3.6 & 9.7 & 8.8 & 12.9 / 25.9 &5.1 / 9.4 &4.7 / 3.1   \\
& SAMV-DUSt3R$_{\text{unrank}}$   & 6.6  & 6.4 & 2.3 & 6.5 & 6.8 & 13.2 / 27.9 &5.2 / 9.5 &4.6 / 2.3  \\
& Ours           & \textbf{6.1}  & \textbf{6.9} & \textbf{1.7} & \textbf{4.7} & \textbf{4.9} & \textbf{14.9} / \textbf{28.5} &5.6 / \textbf{9.6} & \textbf{4.1} / 1.9   \\
\hline
\end{tabular}
\begin{tablenotes}   
\small
\item Note: Results on DTU \cite{aanaes2016large} dataset with 4/12/24 input views from our model and other DUSt3R-based methods. \textbf{Left:} To facilitate metric comparison, we scaled ND down by 10, while scaling DAc, CD, Acc, and Comp up by 10. The results of Multi-View Stereo Reconstruction demonstrate the superior performance of our method across varying numbers of input views. \textbf{Right:} We scale up metrics SSIM, LPIPS and $S_{msk}$ by 10× and $L_{msk}$ by 100000×. We find our method does not always have a good performance, because the test-image and the rendered image incorporate all content from the input views. We exclusively use the disentangled target as the initial point cloud for 3DGS training so the scene content beyond the target region introduces reconstruction errors.
\end{tablenotes}
\end{threeparttable}
\end{table*}

\subsection{Loss function}
To address distinct improvement targets, we trained two dedicated models: a Spatial RankGNN for reference view selection and a Cross Flow Mask Block for enhanced point map regression.
\textbf{Score Rank Loss.} We processed 3600 multi-view scenes using SfM pipeline. The number of images per scene was categorized into 4/12/24 views, with a corresponding distribution ratio of 5:8:23. Each scene is from a co-visibility graph structure, where an image represents a node \textbf{V}, and the number of covisible points between any two images defines the edge weight \textbf{E}. The resulting dataset provides a set of views alongside thier scores. These scores quantify the relative amount of spatial information contributed by each individual view. The GT score is computed as the sum of covisible points between adjacent nodes. These sum values are then normalized across all image nodes using Min-Max scaling to the range [0, 1]. Similarly, the scores predicted by the Spatial RankGNN model are also subjected to the same [0, 1] range. The model is trained using a \textbf{Score Rank Loss}. Crucially, the scores for images are defined relative to their specific view set composition. Our primary objective is to learn the ranking of views based on their spatial information content, rather than precisely regressing their absolute score magnitudes. Consequently, the loss function should prioritize the model's ability to produce correct ordinal rankings. The Score Rank Loss combines MSE loss and Margin Ranking Loss, where the weighting factor $\alpha$ is set to 0.2. \begin{equation}
\textit{L}_{scorerank} = \alpha \cdot \textit{L}_{mse}  + (1 - \alpha) \cdot \textit{L}_{marginranking}
\end{equation}
The Spatial RankGNN module was trained for 5 hours on 8 × NVIDIA A800 GPUs. A prediction is considered correct if the difference between the predicted rank and the GT rank for a view falls within the range [-1, 0, 1] (except for scenes containing only 4 views, where this tolerance criterion was not applied). The Spatial RankGNN model achieves a final ranking accuracy of 73.5\%.

\textbf{Masked Confidence Loss.} We leverage \textbf{(X, C, M)} for pixel-wise 3D object decoupling. Our method targets cognitively distinct objects that are neither visually prominent nor obscure. Such objects are deemed worthy of decoupling from sparse views and must appear in most input views. Once a prompt-specified target is confirmed decoupling-worthy, its mask M within each triplet directly identifies pixels relevant for reconstruction. Given the identical spatial dimensions H×W of \textbf{X}, \textbf{C}, and \textbf{M}, isolating the target region is one-step operation and yields a decoupled duplet \textbf{$(X_m, C_m)$} focused exclusively on the target object. Critical Insight: While \textbf{X} contains regression data for 3D reconstruction, directly using it is insufficient. Decoupling-worthy objects' pixels exhibit bad reconstruction. These pixels suffer from critically low confidence, indicating unreliable 3D position regression. We term this phenomenon the “confidence basin” problem. Addressing the Confidence Basin: Naïvely boosting confidence for basin pixels would not improve reconstruction quality during the model inferencing, as their underlying 3D positions remain poorly regressed. Inspired by DUSt3R’s confidence-aware loss, which jointly enhances confidence estimation and 3D position accuracy, we design a novel target-sensitive confidence loss \textbf{Masked Confidence Loss}, which can specifically optimizes confidence distributions within target mask regions.
We retain the confidence loss $\mathcal{L}_{\text{conf}}$ employed in DUSt3R-based models, which adopts confidence-weighted regression error with a regularization term:
\begin{equation}\mathcal{L}_{\text{conf}} = \sum_{v \in \{1, \ldots, N\}} \sum_{p \in P^v} C_p^{v,r} \ell_{\text{regr}}(v, p) - \beta \log C_p^{v,r}.
\end{equation}
Building upon this, we define a binary foreground mask $M_p^v\in{0,1}$ where $M_p^v=1$ indicates pixel $p$ resides within the foreground region. To prioritize foreground pixels in the loss calculation, we introduce the weighting factor
\begin{equation}
w_p^v=1+\gamma M_p^v (\gamma=0.2),
\end{equation}
amplifying regression errors for masked pixels while maintaining baseline weights for background regions. The regression loss thus becomes $C_p^{v,r} w_p^v \ell_{\text{regr}}(v,p)$, assigning higher weight to masked pixels without neglecting background points entirely. Furthermore, to reshape the confidence distribution within target regions and prevent confidence degradation, we incorporate a uniformity regularizer: we compute the average masked confidence per view $v$ as $\bar{\textbf{C}}^v$ firstly, to encourage homogeneous high-confidence distributions within masked areas, then add the regularizer into $\mathcal{L}_{\text{conf}}$ with $\lambda = 0.01$:
\begin{equation}
\mathcal{L}_{\text{conf}}^{\text{msk}} = \mathcal{L}_{\text{conf}} + \lambda \sum_{v \in \{1, \ldots, N\}}\sum_{p\in P^v}M_p^v(C_p^{v,r}-\bar{C}^v)^2.
\end{equation}
This term minimizes intra-mask confidence variance, promoting uniformity among $C_p^{v,r}$ values. Crucially, the original $-\beta\log C_p^{v,r}$ term persists, ensuring this regularization elevates rather than suppresses confidence, thereby encouraging uniformly high confidence within masked regions.
\begin{figure}[t]
\centering
\includegraphics[width=0.47\textwidth]{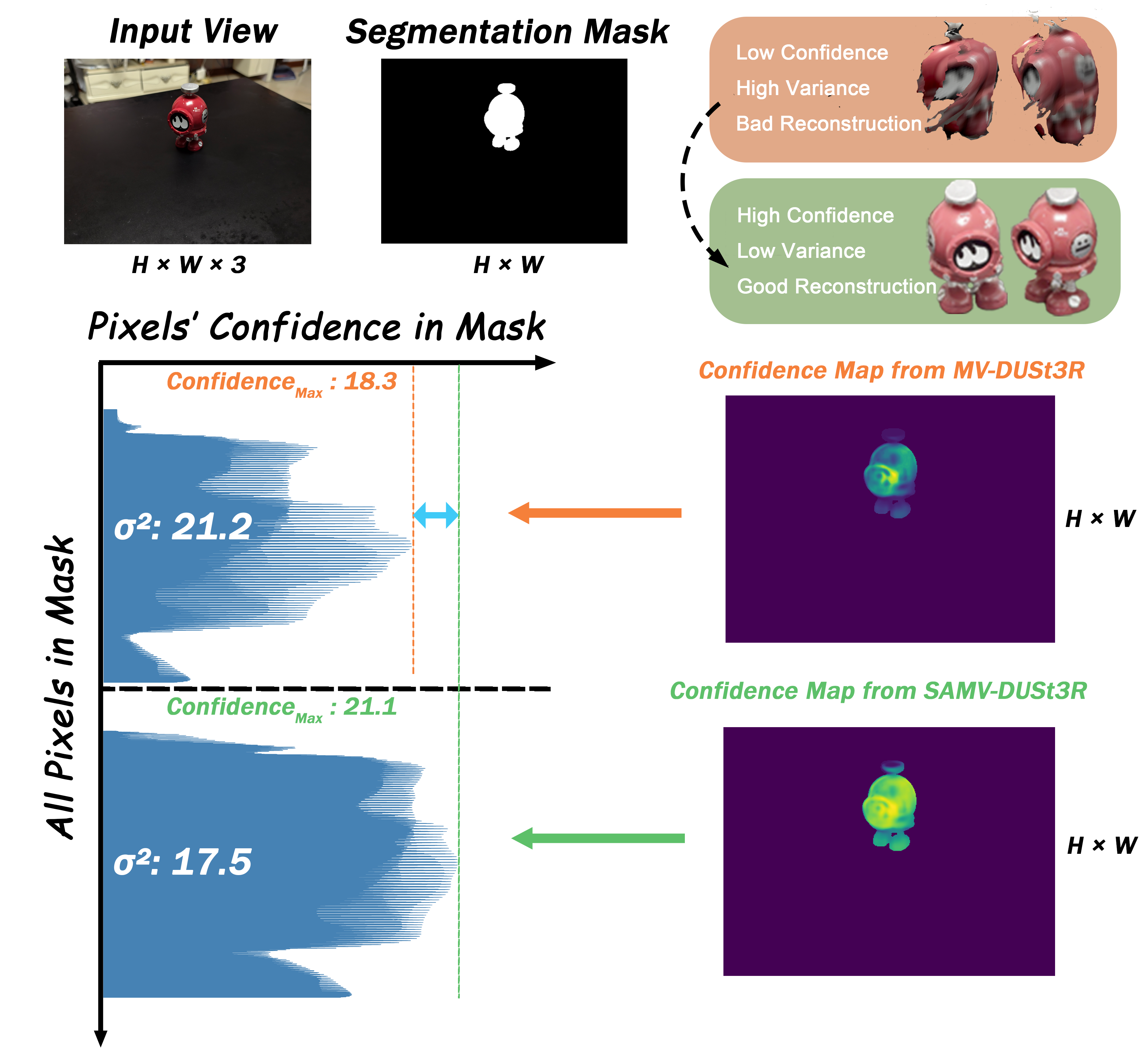} 
\caption{\textbf{Left}: Per-pixel confidence distribution within the masked region, including maximum confidence and confidence variance.\textbf{Top-right}: Comparative visualization of target disentanglement effects under the same scene.\textbf{Bottom-right}: Single-channel confidence heatmap visualization for the same target.}
\vspace{-15pt}
\label{fig:confidence_change}
\end{figure}
\vspace{-5pt}
\section{Experiments}
\subsection{Datasets}
As detailed in Section 3.3, we train the Spatial RankGNN model using a custom-built dataset. Through a SfM pipeline, we construct input 4/12/24 views per scene, distributed in 5:8:23 across 3,600 scenes. Each scene is represented by a co-visibility graph structure, where the input views constitute the nodes, and the number of co-visible points between view pairs defines the edge weights.

For training the Cross Flow Mask Block, we employ object-centric 3D datasets: Co3Dv2 \cite{reizenstein2021commonobjects3dlargescale} and WildRGBD \cite{xia2024rgbdobjectswildscaling} . Both datasets provide RGB images, depth maps, camera parameters and GT masks required for this study. Finally, we evaluate SAMV-DUSt3R's disentanglement capability using the object-centric DTU dataset.

\subsection{Multi-View Stereo Reconstruction}
\subsubsection{Metrics} We measure Normal Deviation (ND), Directional Accuracy (DAc), and Chamfer Distance (CD) to assess local surface detail and normal precision, alongside accuracy (Acc) and completeness (Comp) to quantify global reconstruction fidelity and coverage.
\subsubsection{Details} Each scene of DTU  contains 49 views, we uniformly sample the input views by setting the step size to ensure even distribution. Each view is resized to 224×224. To maximize retention of object regions, the cropping window is centered on the mask area. The corresponding quantitative results are presented in Table 1.

\subsection{Novel View Synthesis}
\subsubsection{Metrics} We compute Peak Signal‑to‑Noise Ratio (PSNR), Structural Similarity Index (SSIM), and Learned Perceptual Image Patch Similarity (LPIPS) to quantify the disparity between rendered views and their corresponding ground‑truth images.
\subsubsection{Details} Using the pointmaps masked by object area on Co3Dv2 and WildRGBD as initial point clouds, we initialize the positions of 3D Gaussians using the points from the predicted pointmaps. Novel views are rendered using camera parameters outside the input views. To minimize the impact of background on metric computation, we zero out pixels outside the mask region in both GT and rendered images to calculate the masked metrics: $ P_{msk} $, $ S_{msk} $ and $ L_{msk} $.

\subsection{Ablation Study}
\subsubsection{Cross Flow Mask Block} MV-DUSt3R was not trained on object-centric datasets and the Cross Flow Mask Block requires object masks, conducting an ablation study directly using MV-DUSt3R's predictions is invalid. Therefore, we train MV-DUSt3R on Co3Dv2 and WildRGBD using the Masked Confidence Loss to get MV-DUSt3R$_{\text{object}}$, which acquires fundamental object-centric reconstruction capability. Subsequently, we train SAMV-DUSt3R following the same procedure. We derive SAMV-DUSt3R$_{\text{unrank}}$ by disabling the Spatial RankGNN module. For both models MV-DUSt3R$_{\text{object}}$ and SAMV-DUSt3R$_{\text{unrank}}$, the first input view is used as the default reference view. The comparative results of these models are presented in Table 1.

\subsubsection{Spatial RankGNN} Our Spatial RankGNN operates solely on the image embeddings generated by the reconstruction network's encoder. Its only outputs ranking scores without modifying the image embeddings themselves. So we can directly integrate the Spatial RankGNN module into SAMV-DUSt3R$_{\text{unrank}}$, yielding the complete SAMV-DUSt3R (Ours). As demonstrated in Table 1, the Spatial RankGNN significantly enhances reconstruction accuracy by automatically selecting the most suitable reference view.

\section{Conclusion}
We propose SAMV-DUSt3R, decoupling 3D objects within a scene from unordered and uncalibrated RGB views, guided by user prompts. SAMV-DUSt3R extends the 2D segmentation to the pointmaps generated by the MV-DUSt3R reconstruction network. Concurrently, we designed the Spatial RankGNN to automate reference view selection to mitigate the impact of manual selection. Furthermore, we introduce the Cross Flow Mask Block to integrate mask of the target into the reconstruction process, guiding the model to focus more effectively on regressing points within the target. Extensive experimentation demonstrates that SAMV-DUSt3R can efficiently disentangle 3D target objects within scenes, offering a potent tool and unlocking new possibilities for downstream tasks in the field of 3D vision.

\bibliographystyle{IEEEbib}
\bibliography{icme2025references}

\end{document}